%% file: camera.tex
\documentclass[letterpaper]{article} 
\usepackage{aaai2026} 
\usepackage{times}  
\usepackage{helvet}  
\usepackage{courier}  
\usepackage[hyphens]{url}  
\usepackage{graphicx} 
\usepackage{natbib}  
\usepackage{caption} 
\usepackage{algorithm}
\usepackage{algorithmic}
\usepackage{booktabs}        
\usepackage{amsmath}
\usepackage{multirow} 
\usepackage[table]{xcolor}  
\usepackage{colortbl}   

\usepackage[utf8]{inputenc} 
\usepackage[T1]{fontenc}    
\usepackage{url}            
\usepackage{booktabs}       
\usepackage{amsfonts}       
\usepackage{nicefrac}       
\usepackage{microtype}      
\usepackage{xcolor}         

\definecolor{darkRed}{RGB}{180,0,0}

\usepackage{bbding}   
\usepackage{pifont}

\usepackage{newfloat}
\usepackage{listings}
\DeclareCaptionStyle{ruled}{labelfont=normalfont,labelsep=colon,strut=off} 
\floatstyle{ruled}
\newfloat{listing}{tb}{lst}{}
\floatname{listing}{Listing}
\usepackage[export]{adjustbox} 

\title{VideoChat-A1: Thinking with Long Videos by Chain-of-Shot Reasoning}

\author{
\begin{tabular}{c}
Zikang Wang\textsuperscript{\rm1,\rm2,\thanks{Equal contribution.}},Boyu Chen\textsuperscript{\rm3,\rm4,\rm6,\footnotemark[1]},Zhengrong Yue\textsuperscript{\rm1,\rm2,\footnotemark[1]},Yi Wang\textsuperscript{\rm2},Yu Qiao\textsuperscript{\rm2},Limin Wang\textsuperscript{\rm5,\rm2},Yali Wang\textsuperscript{\rm3,\rm2,\thanks{ Corresponding author.}}
\end{tabular}
}
\affiliations{
\textsuperscript{\rm1}Shanghai Jiao Tong University  \\
\textsuperscript{\rm2}Shanghai Artificial Intelligence Laboratory\\
\textsuperscript{\rm3}Shenzhen Key Lab of Computer Vision and Pattern Recognition, Shenzhen Institutes of Advanced Technology, Chinese Academy of Sciences\\
\textsuperscript{\rm4}School of Artificial Intelligence, University of Chinese Academy of Sciences \\
\textsuperscript{\rm5}Nanjing University \\
\textsuperscript{\rm6}VIVO AI Lab
}

\usepackage{bibentry}

\begin{document}

\maketitle

\begin{abstract}

Recent advances in video understanding have been driven by MLLMs. 
But these MLLMs are good at analyzing short videos,
while suffering from difficulties in understanding videos with a longer context.
To address this difficulty,
several agent methods have been proposed, 
using MLLMs as agents for retrieving extra contextual knowledge in a long video.
However,
most existing agents ignore the key fact that a long video is composed with multiple shots,
i.e.,
to answer the user question from a long video, 
it is critical to deeply understand its relevant shots like human.
Without such insight,
these agents often mistakenly find redundant even noisy temporal context,
restricting their capacity for long video understanding.
To fill this gap,
we propose VideoChat-A1, 
a novel long video agent paradigm.
Different from the previous works,
our VideoChat-A1 can deeply think with long videos,
via a distinct chain-of-shot reasoning paradigm.
More specifically,
it can progressively select the relevant shots of user question,
and 
look into these shots in a coarse-to-fine partition.
By multi-modal reasoning along the shot chain,
VideoChat-A1 can effectively mimic step-by-step human thinking process,
allowing the interactive discovery of preferable temporal context for thoughtful understanding in long videos.
Extensive experiments show that,
VideoChat-A1 achieves the state-of-the-art performance on the mainstream long video QA benchmarks,
e.g., 
it achieves 77.0 on VideoMME~(w/ subs) and 70.1 on EgoSchema, 
outperforming its strong baselines
(e.g., InternVL2.5-8B and InternVideo2.5-8B),
by up to 10.1\% and 6.2\%.  
Compared to leading closed-source GPT-4o and Gemini 1.5 Pro,  VideoChat-A1 offers competitive accuracy, 
but only with 7\% input frames and 12\% inference time on average. The code is available on https://github.com/SpXace/VideoChat-A1.
\end{abstract}

\section{Introduction}

Video understanding is an important problem in computer vision~\cite{you2024longvideounderstandingfinedetailed}.
With fast development of MLLMs,
video understanding has achieved significant progress~\cite{gpt4o,wang2025internvideo25}.
However,
most existing MLLMs are good at understanding short videos in seconds,
while hardly analyzing longer videos in minutes (or hours) properly.
The main challenge lies in feeding a huge amount of frames in long videos to MLLMs.
To deal with this problem,
several attempts have been proposed by 
modeling such long multimodal context~\cite{gpt4o, largeworldmodel}, 
or 
equipping MLLMs with token compression~\cite{videochatflash, videoxl}, 
But their efficiency and effectiveness still need to be further improved in tackling redundant video content,
thus blocking their performance on long video understanding benchmarks.

\begin{figure*}[t]     
  \centering       
  \includegraphics[width=\textwidth]{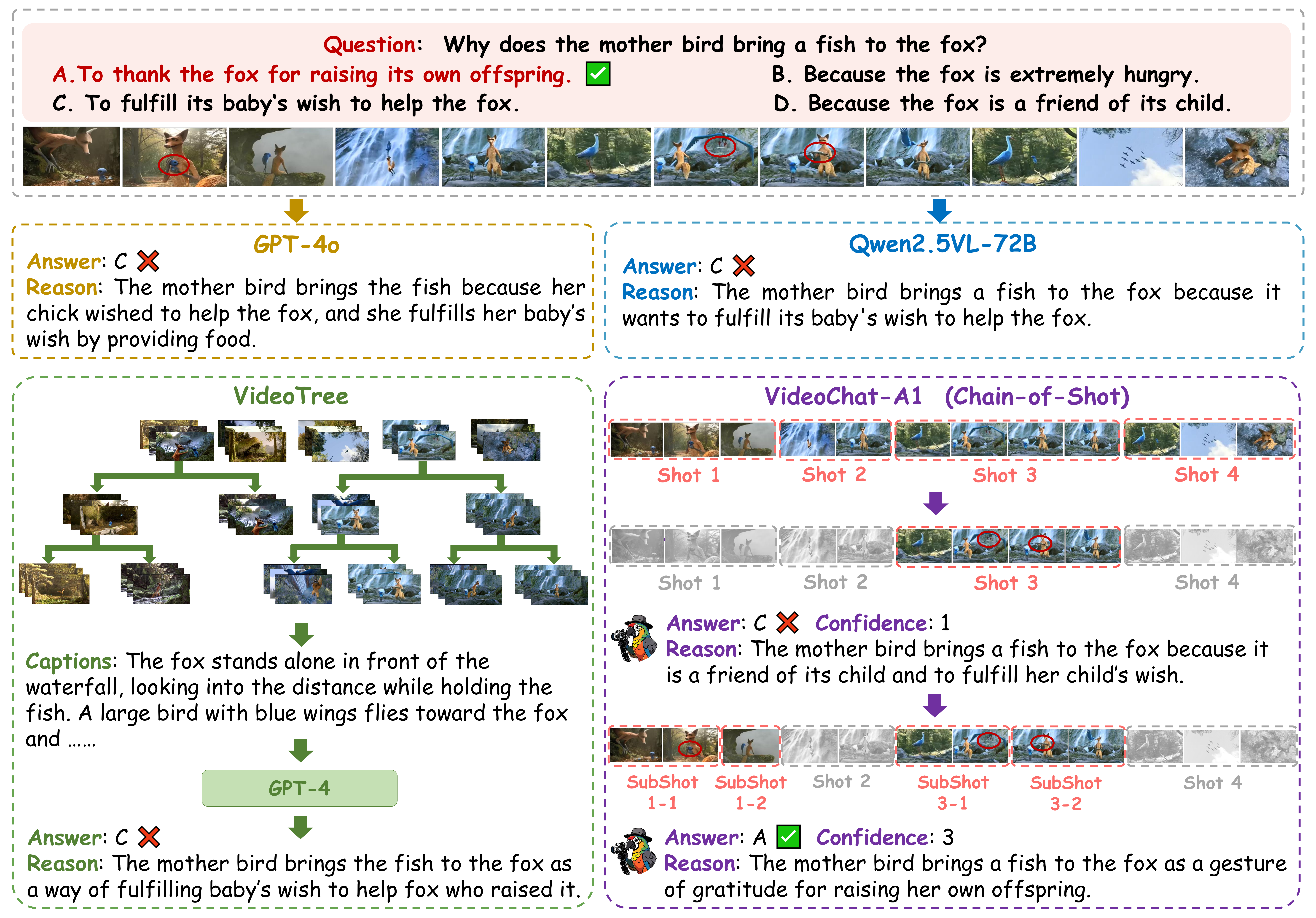}
  \caption{\textbf{Motivation.} Direct reasoning models such as GPT-4o~\cite{gpt4o}  perform global sampling and struggle to focus on key information. Agent-based methods like VideoTree~\cite{wang2024videotree} often suffer from incorrect or redundant sampling that leads to noisy captions.
  In contrast, VideoChat-A1 interactively employs shot perception and reasoning via Chain-of-Shot, which progressively looks into relevant shots through a reflective process for better performance. }
  \label{fig:visualization}  
\end{figure*}

Recent studies have shown that,
agent-based approaches are promising for long video understanding ~\cite{wang2024videoagent, chen2025lvagent, videomultiagents, videoagent2,wang2024videotree}. 
Instead of feeding the entire long video with model adaptation,
they simplify this challenge by invoking various tools to extract and retrieve relevant information from long videos.
However,
these approaches lack flexibility to capture complex content changes in a long video,
since the retrieval is based on off-the-shelf video knowledge which is fixed after extraction.
Alternatively,
OpenAI o3~\cite{o3} shows a human-like thinking process on images.
Via chain-of-thought reasoning,
o3 can deeply understand visual content via progressive interaction with images.
However,
they ignore that,
a long video consists of multiple shots.
To answer the user question reliably,
it is critical to deeply understand its relevant shots via multi-round reasoning.
Without this consideration,
these agents often mistakenly find redundant even noisy temporal context, 
restricting their capacity for long video understanding.

To fill this gap,
we propose VideoChat-A1,
a novel video agent framework that can progressively think with a long video,
via an interactive chain-of-shot reasoning paradigm.
Specifically,
for each shot reasoning step,
we leverage MLLM as core agent of thinking,
and invoke tools for selecting relevant shots,
dividing these shots into subshots,
and 
reasoning answer with subshots.
If the answer is unconfident,
it means that the current subshots are insufficient for understanding user question.
As a result,
VideoChat-A1 will start next-step shot reasoning to further understand user question 
with finer relevant shots.
Through such a distinct chain-of-shot paradigm,
our VideoChat-A1 can effectively mimic the thinking process of human,
by progressively reasoning on user answer while iteratively looking into relevant video shots, as shown in Fig.~\ref{fig:visualization}.
Extensive experiments show that, 
our VideoChat-A1 achieves the state-of-the-art performance on 4 mainstream benchmarks,
e.g., 
it achieves 77.0 on VideoMME with subset and 70.1 on EgoSchema, 
outperforming its baselines such as  
InternVL2.5-8B 
and 
InternVideo2.5-8B,
by up to 10.1\% and 6.2\%.
Compared to GPT-4o and Gemini 1.5 Pro, 
VideoChat-A1 offers competitive accuracy, 
but with 7\% input frames and 12\% inference time on average.

\section{Related Work}

\paragraph{MLLM for Long Video Understanding.}
MLLMs have demonstrated great potential in the field of video understanding~\cite{gpt4o,videochat,chen2022low}. 
However, they still face challenges when dealing with long videos that can be minutes or hours in length
~\cite{moviechat}.
Currently, there are two main approaches to addressing this problem. One is to compress visual tokens~\cite{moviechat, timesuite}. For example, MA-LLM~\cite{malmm} merges similar tokens, and LLaMA-VID~\cite{li2025llama} compresses each frame into context and content tokens to enable longer video input. However, these methods result in a significant loss of visual information. 
Another approach is to extend the number of processable tokens~\cite{lwm,longllava}. Models like LongVILA~\cite{longvila,videochatflash} have adopted strategies to increase the token capacity for handling longer videos. However, this method introduces  redundant information, consumes more computational resources, and often achieves only moderate effectiveness.

\paragraph{Agent Based Method for Long-context Understanding.}
Agent mechanisms~\cite{chen2025lvagent,videomultiagents,ye2025rethinking} have been introduced in video understanding tasks. 
Currently, Agent methods for long video understanding can be mainly divided into two categories.
One is extracting knowledge from long videos by invoking external tools at once and answering questions by retrieving the extracted knowledge.
For example, CLIP~\cite{clip} is used for retrieving key frames related to the questions~\cite{zhao2024longagent,xie2024openagents}. Memory banks~\cite{videoagentmemory,malmm} and search engines~\cite{searchlvlm,chen2024sharegpt4video} are applied for extracting key information of the videos.
However, this single-pass information extraction often misses critical details   or  introduces redundant information, which interferes with the model’s response.
The other approach involves answering questions through a round by round key information search. The multi-round pipeline emphasizes active video exploration and dynamic information acquisition~\cite{wang2024videotree,wang2024videoagent,drvideo}, and optimizes the reasoning path through retrieval and reflection mechanism. 
However, a critical limitation is that most  methods either ignore video shots—the fundamental structural units of long-form videos—or segment shots at once without further fine-grained partitioning, which can both miss critical query-relevant details within each shot and introduce redundant or noisy information.
 For instance, VideoINSTA~\cite{liao2024videoinsta} proposes an event-based spatial-temporal reasoning framework, but its spatial and temporal information extraction processes are relatively independent and lack query-driven interactive iterative refinement.

\begin{figure*}[!t]     
  \centering       
  \includegraphics[width=\textwidth]{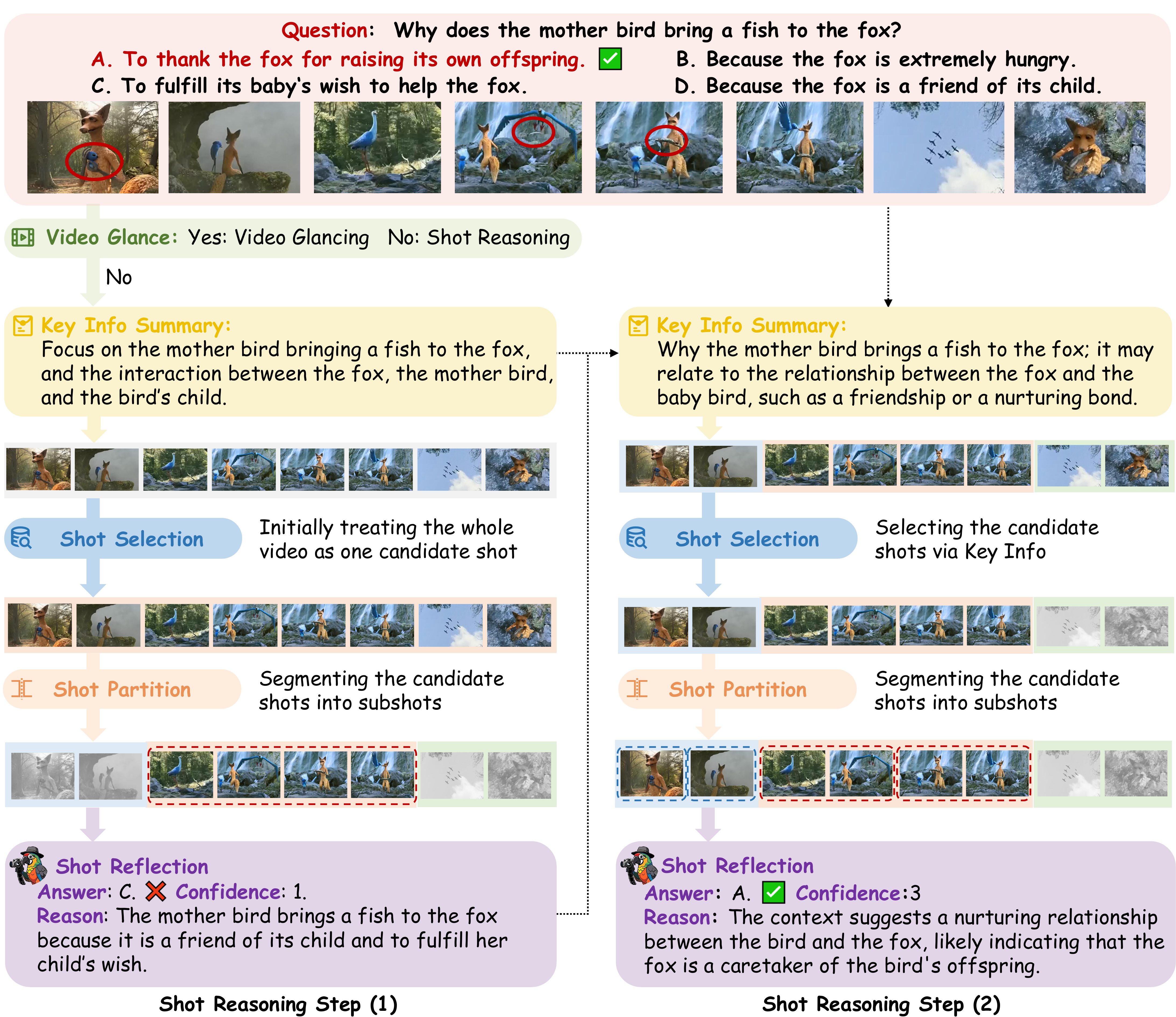}
  \caption{\textbf{Framework.}
  VideoChat-A1 introduces a novel \textbf{Chain-of-Shot Reasoning} framework for long video understanding. It progressively refines video analysis through iterative stages of \textit{Shot Selection}, \textit{Shot Partition}, and \textit{Shot Reflection}, leveraging MLLMs to dynamically discover relevant video shots and generate reliable answer.}
  \label{fig:framework}  
\end{figure*}

\section{Method}

In this section,
we introduce our VideoChat-A1.
To deeply understand the user question in a long video,
we design a distinct \textbf{Chain-of-Shot (CoS)} reasoning paradigm,
which can progressively think while discovering relevant shots
via multi-round shot partition and dialogues. 
Specifically,
it consists of three key steps in each round of shot reasoning,
including \textbf{Shot Selection}, \textbf{Shot Partition}, and \textbf{Shot Reflection}. The framework of VideoChat-A1 is shown in Fig.~\ref{fig:framework}.

\textbf{Video Glance}.
To start with,
it is necessary to determine if the question of the given video needs to dig into the local shots for answering,
since some questions may refer to the global content throughout the entire video.
In this regard,
we introduce a concise video glance step for pre-judging.
First,
we uniformly sample 4 frames from the entire video to roughly describe the video.
Then,
we feed these frames along with user question and options into MLLM,
justifying whether it is necessary to view the entire video to answer this question.
If MLLM considers the question to be global,
we uniformly sample 32 frames from the entire video,
and feed them into MLLM for answering.
If MLLM considers the question to be local,
we start chain-of-shot reasoning.

\subsection{Shot Selection}

For conciseness,
we describe the $i$-th shot reasoning step for illustration.
Suppose that,
we divide the whole video into $M$ shots at the $(i-1)$-th round,
\begin{equation}
\mathcal{S}^{(i-1)}=\{\mathcal{S}^{(i-1)}_{1},...,\mathcal{S}^{(i-1)}_{m},...,\mathcal{S}^{(i-1)}_{M}\}, \label{eq:shot}
\end{equation}
As shown in Fig.~\ref{fig:framework},
we first select the candidate shots from the existing ones at the $(i-1)$-th round,
in order to further look into these candidates for reasoning at the $i$-th round.

\textbf{Key Information Summary}.
To achieve this goal,
we leverage MLLM to summarize key text information $\mathcal{I}^{(i)}$ to describe how to answer the user question,
according to the reasoning results in the previous round,
\begin{equation}
\mathcal{I}^{(i)}=\text{MLLM}(\mathcal{Q},~\mathcal{O},~\mathcal{V}^{(i-1)},~\mathcal{H}^{(i-1)}),\label{eq:keyinfo}
\end{equation}
where
$\mathcal{Q}$ and $\mathcal{O}$ refers to the user question and answer options.
$\mathcal{V}^{(i-1)}$ refers to the video frames sampled from the shots in the historical rounds.
Moreover,
$\mathcal{H}^{(i-1)}$ refers to historical reasoning information,
including 
key text information $\mathcal{I}^{(i-1)}$,
the chosen answer $\mathcal{A}^{(i-1)}$,
and 
the reason why to choose this answer $\mathcal{R}^{(i-1)}$ at round $(i-1)$.

\textbf{Shot Selection via Retrieval}.
After obtaining the key text information $\mathcal{I}^{(i)}$,
we leverage it as contextual guidance
 for shot selection.
This can be effectively achieved by shot-text retrieval,
\begin{equation}
\mathcal{C}^{(i)}=\text{CLIP}(\mathcal{I}^{(i)},~\mathcal{S}^{(i-1)}),\label{eq:shotselect}
\end{equation}
where we leverage our finetuned LongCLIP~\cite{zhang2024longclip} as a retriever, and compute the cosine similarities between key information and each shot in $\mathcal{S}^{(i-1)}$.
Finally,
we select top $N$ shots as candidates for further investigation,
$\mathcal{C}^{(i)}=\{\mathcal{C}_{1}^{(i)},...\mathcal{C}_{N}^{(i)}\}$,
where
$\mathcal{C}_{n}^{(i)}$ refers to the $n$-th selected shot from the shot set $\mathcal{S}^{(i-1)}$.
Additionally,
since there is one single shot (i.e., the whole video) at the 1st round,
we directly use this shot as the candidate at this round.

\subsection{Shot Partition}
\begin{figure*}[t]     
  \centering       
  \includegraphics[width=\textwidth]{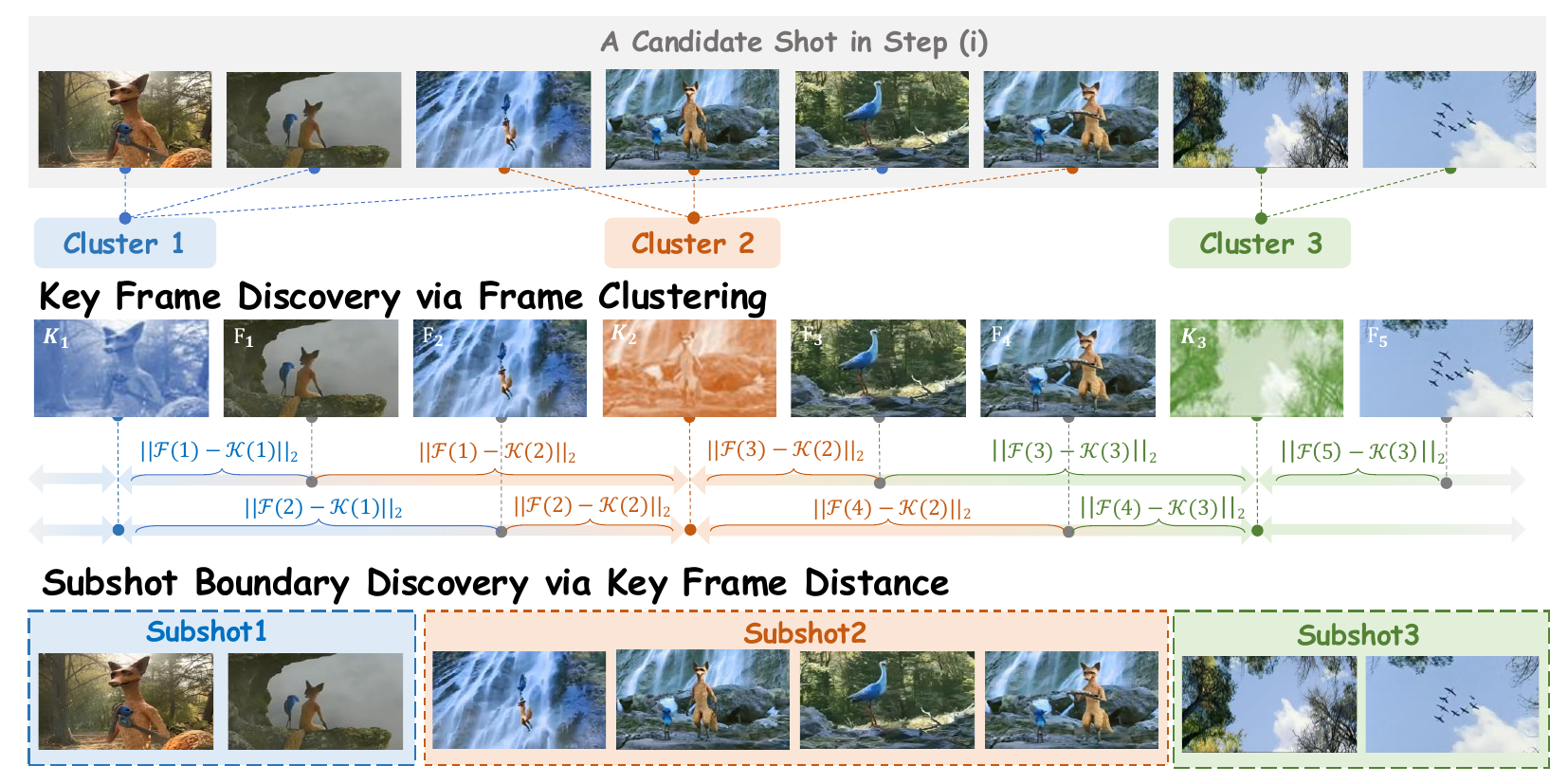}
  \caption{\textbf{Shot Partition.} Given a candidate shot at step $i$, VideoChat-A1 first applies K-Means to obtain K cluster centers for finding key frames. Subsequently, subshots are partitioned based on the feature distance  and its adjacent two key frames.}
  \label{fig:subshot}  
\end{figure*}

After obtaining the candidate shots $\mathcal{C}^{(i)}$,
we next further divide each of them into subshots, to look deeper into it for reasoning.
In this paper,
we introduce a concise shot partition pipeline as shown in Fig.~\ref{fig:subshot}.
For notation simplicity,
we illustrate how to perform it on a candidate shot.
 
\textbf{Key Frame Discovery}.
To discover the subshots in a candidate shot,
we start by finding  key frames in the subshots.
First,
we uniformly sample frames from the candidate shot.
To reduce temporal redundancy while maintaining computation efficiency,
we sample 1 fps in this paper.
Second,
we extract the features of the sampled frames by CLIP. 
Third,
we perform K-Means~\cite{kmeans} on these features.
As a result,
each of $K$ clusters roughly reflects a subshot. The choice of $K$ is discussed in detail in the ablation study section.
However,
frames in each cluster may not be temporally adjacent,
since K-Means Clustering does not take temporal order into account.
To construct subshots in temporal order,
we further find a key frame in each cluster to divide the shot.
Specifically,
for each cluster,
we choose the frame whose feature is closest to the cluster center,
as the key frame in this cluster.
Finally,
we organize $K$ key frames in the temporal order, 
$\mathcal{K}=\{\mathcal{K}(1),...,\mathcal{K}(K)\}$,
for the candidate shot.

\textbf{Subshot Boundary Discovery}.
After finding $K$ key frames,
we next identify the boundary frame between two adjacent key frames to construct $K$ subshots.
Specifically,
we compute $\mathcal{L}_2$ feature distances between each frame 
$\mathcal{F}$ 
and its two adjacent key frames.
Then we sum over them as a semantic deviation metric for this frame,
\begin{equation}
d=||\mathcal{F}-\mathcal{K}(k)||_2+||\mathcal{F}- \mathcal{K}(k+1)||_2,
\label{eq:distance}
\end{equation}
A higher value of $d$ indicates that this frame lies far from both adjacent key frames, 
implying a semantic shift between two subshots. 
Therefore, 
we select the frame with the maximum $d$ as the boundary frame between two subshots.
With such a mechanism,
each candidate shot is divided into $K$ subshots,
$\mathcal{C}_{n}^{(i)}=\{\mathcal{S}_{n}^{(i)}(1),...,\mathcal{S}_{n}^{(i)}(K)\}$.
Finally,
we update the whole shot set by replacing all the candidate shots as the corresponding subshots,
\begin{equation}
\mathcal{S}^{(i)} \leftarrow 
\left(\mathcal{S}^{(i-1)} \setminus \mathcal{C}^{(i)} \right)
~\cup~
\{\mathcal{S}_{n}^{(i)}(1),...,\mathcal{S}_{n}^{(i)}(K)\}_{n=1}^{N},
\label{eq:update_shots}
\end{equation}
which is used for shot selection in the next step if necessary.

\subsection{Shot Reflection}

\begin{figure}[!t]
  \centering
  \includegraphics[width=\columnwidth]{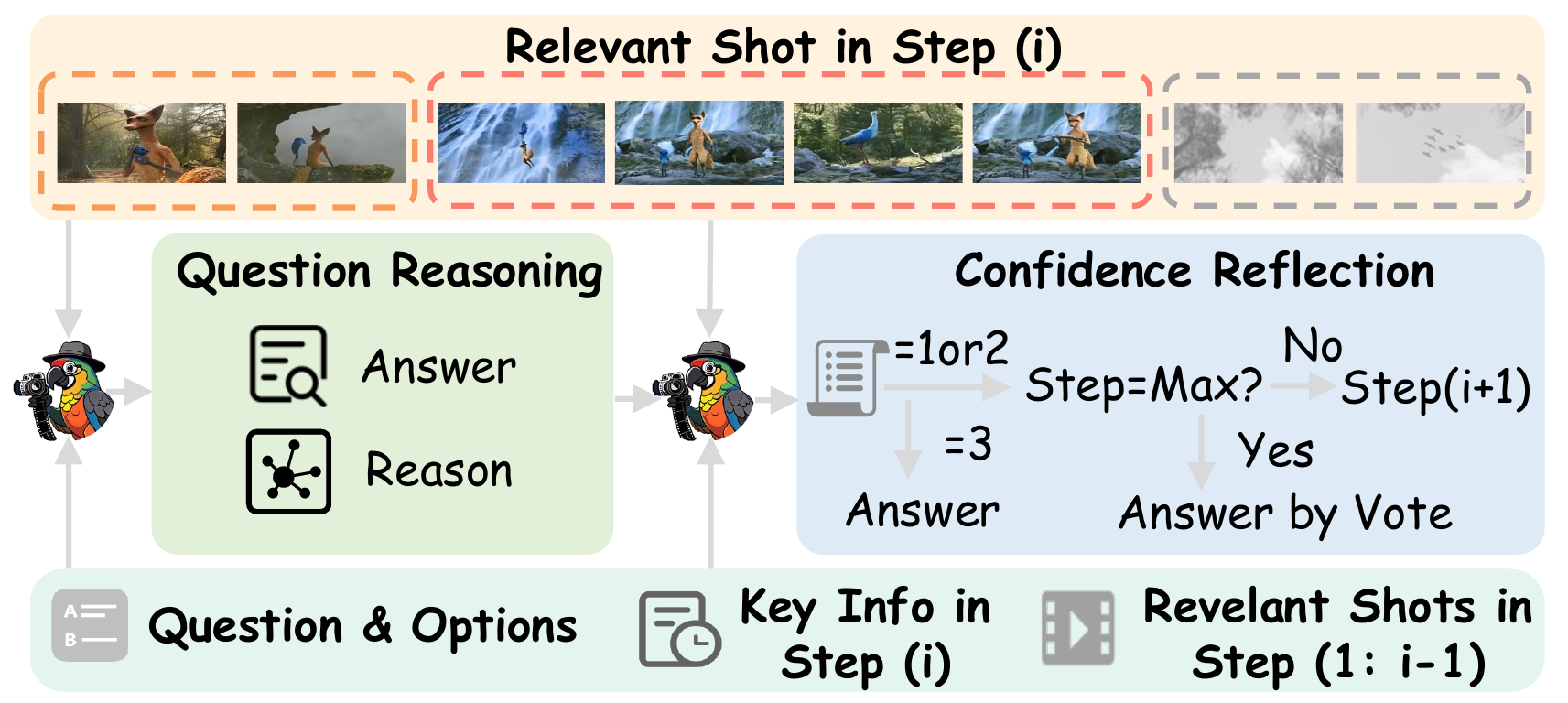}
  \caption{\textbf{Shot Reasoning and Shot Reflection.} 
    At each round, VideoChat-A1 performs question-answering reasoning using the relevant shots identified. It then evaluates 
    the confidence level of the response. 
    Then, the system either proceeds to the next step for further refinement or terminates the reasoning process to output the final answer. 
  }
  \label{fig:reasoning}   
\end{figure}

\textbf{Question Reasoning}.
After obtaining subshots from the candidate shots,
we employ MLLMs to answer the user question based on these finer shot regions.
Specifically, as shown in Fig.~\ref{fig:reasoning},
we leverage MLLM as a dialogue agent  to answer the question at round $i$,
\begin{equation}
\{\mathcal{A}^{(i)}, \mathcal{R}^{(i)}\}=\text{MLLM}(\mathcal{Q},~\mathcal{O},~\mathcal{V}^{(i)}),\label{eq:ans}
\end{equation}
where
$\mathcal{Q}$ and $\mathcal{O}$ refer to question and answer options.
We sample $\mathcal{T}$ frames from each subshot $\mathcal{S}_{n}^{(i)}(k)$, where $\mathcal{T}$ is the number of sampled frames, 
and 
combine them with $\mathcal{V}^{(i-1)}$ to form 
$\mathcal{V}^{(i)}$.
Finally,
given these inputs, MLLM generates answer $\mathcal{A}^{(i)}$,
and the reason why to choose this answer $\mathcal{R}^{(i)}$.

\textbf{Confidence Reflection}.
Next,
MLLM would justify if the answer was reliable to decide if we need to next-round shot reasoning.
Hence,
we leverage MLLM again to generate the answer confidence,
with extra inputs of answer and reason,
\begin{equation}
    \mathcal{Z}^{(i)}=\text{MLLM}(\mathcal{Q},~\mathcal{O},~\mathcal{V}^{(i)},\mathcal{A}^{(i)}, \mathcal{R}^{(i)}).
\label{eq:confidence}
\end{equation}
We set the confidence level $\mathcal{Z}^{(i)}$ from 0 to 3.
If the confidence is higher than 2,
we believe that it is a reliable answer as the final output.
Otherwise, we believe that it is an unreliable answer  and conduct further investigation.
Hence,
we start shot selection, partition, reflection in the next round.
Additionally,
we set a maximum number of shot reasoning rounds.
If VideoChat-A1 still does not get the confident answer at the max step,
it will vote the major answer as the final one,
according to the answers of all the rounds.
Via  exploit relevant shots in a coarse-to-fine manner,
our VideoChat-A1 allows to leverage a chain of shots for deep thinking on long videos,
and thus progressively and interactively discover the reliable shot context to answer user question.

\input{table/sota_and_time}

\section{Experiment}

\subsection{Experimental Setup}

We evaluate VideoChat-A1 on four long-video question-answering benchmarks: EgoSchema \cite{egoschema}, LongVideoBench \cite{wu2024longvideobench}, MLVU \cite{MLVU}, and Video-MME \cite{videomme}. As baselines, we use Qwen2.5-VL-7B, InternVL2.5-8B, and InternVideo2.5-8B.
In the Feature Extraction stage, we sample the video at 1 fps and use pretrained CLIP-ViT-B/32~\cite{clip} as the feature extractor. In Shot Selection, we sample 16 frames from each shot and use a fine-tuned LongCLIP-B~\cite{zhang2024longclip} to calculate shot-text similarity. The  details are described in the supplementary documentation. We retrieve shots with similarity scores above the 0.8 threshold. In the first round, we treat the whole video as the candidate shot without selection, and sample $\mathcal{T}=16$ frames as initial frames. In each subsequent round, we select top 2 shots, divide each shot into 2 subshots, and sample $\mathcal{T}=8$ frames from each subshot, and add them into frame set. In the Shot Partition stage, the number of clusters $K$ is set to 6 in 1st round and 2 in subsequent rounds.
The maximum number of iterations is 3. All experiments are conducted on 2 A800-80GB GPUs.
    
\subsection{Comparison with SOTA}

\paragraph{Performance and Efficiency Analysis.}

As shown in Table~\ref{tab:results}, we evaluate VideoChat-A1
on four mainstream benchmarks. Specifically, VideoChat-A1 significantly outperforms open-source 7-8B models, comparable to or exceeding that of large open-source or closed-source models. 
For example,
on MLVU, VideoChat-A1 (\texttt{InternVideo2.5-8B}) achieves 76.2\%, outperforming GPT-4o by 11.6\%.  
Table~\ref{tab:efficient} compares the average frame number and inference time. VideoChat-A1 achieves substantial gains in efficiency. For instance, VideoChat-A1 (\texttt{InternVL2.5-8B})  
reduces frame usage by over 90\% and inference time by more than 85\% compared to GPT-4o, Gemini 1.5 Pro, and Qwen2.5-VL-72B. Despite these significant savings, our method maintains competitive performance across benchmarks. 
In contrast to small open-source models, VideoChat-A1 not only delivers significantly better performance but also requires fewer video frames and achieves faster or comparable inference. 

\subsection{Ablation Study}
\input{table/abla_global}
\paragraph{Ablation study of Video Glance.} 
In this study, we conducted an ablation study to verify the impact of the video glance. Table~\ref{tab:abla_global_obv} shows that the model with the video glance achieves higher scores on all the tested tasks. The improvements proving the effectiveness of video glance. 

\paragraph{Ablation on Shot Partition and Chain-of-Shot.} 
In Table~\ref{tab:process}, we evaluate the impact of Shot Partition~(SP) and Chain-of-Shot~(CoS) on VideoChat-A1 (\texttt{Qwen2.5-VL-7B}). The EgoSchema and LongVideoBench are abbreviated as Ego and LVbench in the table because of the limited space. In the setup without SP, we evenly divided the entire video into six shots on average in the 1st round. In subsequent rounds, for the selected shots that need to be expanded, they are directly partitioned into two shots on average.
Regarding the setup without CoS, we removed the CoS mechanism. Instead,  
the model samples additional frames from the video and incorporates them into the existing frame set from the previous round. Then we conducted a vote to get the final response. 
Individually, enabling SP or CoS boosts performance. Their combined use yields the best results. For example, on MLVU, the score rises from 66.2 (with neither) to 71.9 (both enabled).

\input{table/abla_reflect}
\input{table/abla_big}

\paragraph{Ablation on Shot Partition Methods.}

As shown in Table~\ref{tab:shot_partition}, we evaluate the impact of three methods of shot partition across multiple benchmarks. Cluster Partition refers to the clustering approach used in VideoTree~\cite{wang2024videotree}, where all frames within each cluster are treated as a single shot, even though they may not form a continuous video segment. Average Partition refers to dividing each shot into smaller shots by uniformly splitting it into equal parts. Our proposed Shot Partition method consistently outperforms 
the other two methods, which  demonstrate that partitioning videos into temporally coherent shots based on semantic information leads to  better understanding of long videos.

\paragraph{Ablation Max Rounds of Reflection.}
We evaluate the impact of varying the maximum number of reflection rounds. As shown in Table~\ref{tab:abla_maxround}, increasing the number of rounds consistently leads to performance improvements. 
In the context of multi-round discussions, increasing the number of discussion rounds tends to enhance the effectiveness. After three rounds of the chain-of-shot process, the results are already quite satisfactory. As the fourth round yields only marginal improvements, we ultimately adopt the three-round setting to balance performance gains and computational cost.

\paragraph{Ablation on Number of Initialized Shots.} 

We evaluate the effect of the number of initial shots, which is set to 3, 6, and 9. As shown in Table~\ref{tab:num_init_shot}, the 6-shot  achieves best accuracies compared  with the other two configurations.
For the subsequent rounds, considering that iterative process generates many subshots and aiming to balance computational efficiency, we fix  clustering number for each subshot at $K=2$.

\section{Conclusion}

In this paper, we introduced \textbf{VideoChat-A1}, an agent-based framework for effectively addressing long video understanding tasks. Unlike existing methods that often overlook the intrinsic shot-based structure , our method explicitly employs a \textbf{Chain-of-Shot reasoning paradigm}. Specifically, VideoChat-A1 progressively identifies and refines relevant video segments through three key iterative steps: \textbf{Shot Selection}, \textbf{Shot Partition}, and \textbf{Shot Reflection}. This iterative interaction mechanism closely mimics human cognitive processes, allowing the agent to thoughtfully reason and accurately answer the questions. Experimental results demonstrate the effectiveness and efficiency of our method, highlighting the substantial benefits   in long video understanding tasks.

\section{Acknowledgements}
Supported by Shanghai Artificial Intelligence Laboratory, and the National Key R\&D Program of China(NO.2022ZD0160505).

\bibliography{Reference}

\end{document}

%% file: table/sota_and_time.tex
\begingroup
  \setlength{\floatsep}{0pt}
  \setlength{\textfloatsep}{0pt}
  \setlength{\intextsep}{0pt}

\begin{table*}[t]
\centering
\resizebox{\textwidth}{!}{
\begin{tabular}{lccccccc}
\toprule
\small
\multirow{2}{*}{\textbf{Model}}  & \multirow{2}{*}{\textbf{EgoSchema}} & \multirow{2}{*}{\textbf{LongVideoBench}} & \multirow{2}{*}{\textbf{MLVU }} & \multicolumn{4}{c}{\textbf{Video-MME}} \\
\cmidrule(lr){5-8}
&  &  &  \textbf{(M-avg)} & \textbf{Short} & \textbf{Medium} & \textbf{Long} & \textbf{Average}  \\

\midrule
\rowcolor{gray!20} \textit{Closed-Source Model} & & & & & & & \\

GPT-4o(0513)~\cite{gpt4o}  & 72.2 &  {66.7} & 64.6 & 80.0 / 82.8 & 70.3 / 76.6 & 65.3 / 72.1 & 71.9 / 77.2 \\
Gemini 1.5 Pro~\cite{gemini15}  & 71.1 & 64.0 & - & 81.7 /  {84.5} &  {74.3} /  {81.0} & 67.4 /  {77.4} & 75.0 /  {81.3}\\

\midrule
\rowcolor{gray!20} \textit{Open-Source 72B-78B Model} & & & & & & & \\

   LLaVA-OneVision-72B~\cite{llavaonevision}
                               & 62.0 & 63.2 & 68.0 & 76.7/79.3 & 62.2/66.9 & 60.0/62.4 & 66.3/69.6 \\
   VideoLLaMA-2-72B~\cite{videollama2}
                               & 63.9 &  -  & 45.6 & 69.8/72.0 & 59.9/63.0 & 57.6/59.0 & 62.4/64.7 \\
  LLAVA-Video-72B~\cite{llavavideo}
                               & 65.6 & 64.9 &  -  & 81.4/82.8 & 68.9/75.6 & 61.5/72.5 & 70.6/76.9 \\
   Qwen2.5-VL-72B~\cite{qwen25vl}
                               & \,77.9 & -   & -  & 80.1/82.2 & 71.3/76.8 & 62.2/74.3 & 71.2/77.8 \\
   InternVL-2.5-78B~\cite{internvl2.5}
                               &  -  & 63.6 & 75.7 & 82.8/83.2 & 70.9/74.1 & 62.6/64.8 & 72.1/74.0 \\
   InternVL-3-78B~\cite{internvl3}
                               & -  & 65.7 & 79.5 & - & - & - & 72.7 /75.7 \\

\midrule
\rowcolor{gray!20} \textit{Open-Source 7B-8B Model} & & & & & & & \\

 ShareGPT4Video-8B~\cite{chen2024sharegpt4video}
                               &  -  & 39.7 & 46.4 & 48.3/53.6 & 36.3/39.3 & 35.0/37.9 & 39.9/43.6 \\
  VideoChat2-7B~\cite{mvbench}    & 56.7 & 39.3 & 47.9 & 48.3/52.8 & 37.0/39.4 & 33.2/39.2 & 39.5/43.8 \\
  LongVU-7B~\cite{longvu}         & 67.6 &  -  & 65.4 &   -     &  -     & -/59.5   & -/60.6   \\
  LLaVA-Video-7B~\cite{llavavideo}& 57.3 & 58.2 & 70.8 &  -      &  -    & -& 63.3/69.7 \\
  
 Qwen2.5-VL-7B~\cite{qwen25vl}   & 65.0 & 56.0 &  -  &   -    &   -   &   -     & 65.1/71.6 \\

    InternVideo-2.5-8B~\cite{wang2025internvideo25}
                               &  63.9   & 60.6 & 72.8 &  -     &   -    &   -     & 65.1/- \\
   InternVL-2.5-8B~\cite{internvl2.5}
                               & -  & 60.0 & 68.9 &   -     &  -     &   -     & 64.2/66.9 \\
  InternVL-3-8B~\cite{internvl3}
                               &  -   & 58.8 & 71.4 &   -    &    -      &    -      & 66.3/68.9 \\   
\midrule

\rowcolor{gray!20} \textit{Agent Based Model} & & & & & & & \\

VideoAgent~\cite{wang2024videoagent} &54.1 & -& -& - & - & - & - \\
VideoTree~\footnotesize{(\textbf{\texttt{GPT-4}})} ~\cite{wang2024videotree} & 61.1& - & -& - & - & - & - \\
DrVideo~\cite{drvideo}   & 61.0  & -	 & -& - & - & - & 51.7/71.7\\
T*~\cite{ye2025rethinking}  & -  & -	 & -&61.0/-& 66.6/-& 77.5/-& 68.3/-\\
VideoMind~\cite{videomind}  & -  & -	 & 64.4 & -& -&49.2/-& 58.2/-\\
VideoMultiAgents~\cite{videomultiagents}  & 68.0  & -	 & - & -&-& -& -\\
VideoRAG-72B~\cite{videoragvideofeat}   & - & 65.4 & 73.8 & 81.1/- & 72.9/- &73.1/- &  75.7/- \\
\midrule

\textbf{VideoChat-A1}~\footnotesize{(\textbf{\texttt{Qwen2.5-VL-7B}})} & 70.7  & 64.2 & 71.9 &78.0/81.2 & 73.1/77.4 & 64.3/70.9 & 71.8/76.5 \\

\textbf{VideoChat-A1}~\footnotesize{(\textbf{\texttt{InternVL2.5-8B}})} & 72.1  & 65.2 &75.1  & 80.8/81.4 & 72.3/76.8  & 63.7/72.6 & 72.3/77.0 \\

\textbf{VideoChat-A1}~\footnotesize{(\textbf{\texttt{InternVideo2.5-8B}})} & 70.1 & 65.4 & 76.2 & 81.4/82.4 & 72.8/76.7 & 65.0/71.2 &72.9/76.8 \\

\bottomrule
\end{tabular}
}

\caption{Comparison with closed-source, open-source, and agent based model on four Long Video Understanding Tasks. The Video-MME results are presented in the format ``w/o subs / w/ subs".}

\label{tab:results}
  \end{table*}
\begin{table*}[t]{
        \centering
        \small
        \setlength{\tabcolsep}{3pt} 
        \resizebox{\textwidth}{!}{
            \begin{tabular}{c|cccccc}
               
                \toprule
                \textbf{Model} & \textbf{Frames} & \textbf{Inference Time}  
                &\textbf{EgoSchema} & \textbf{LongVideoBench}   &\textbf{MLVU}  & \textbf{VideoMME} \\
                \hline
             GPT-4o\cite{gpt4o}      & 384 &     134.4s     & 72.2    & 66.7 & 64.6 &  71.9/77.2   \\
 
          Gemini-1.5-Pro\cite{gemini15}   & 568 &  198.8s    & 71.1  &  64.0 & - & 75.0/81.3     \\
     Qwen2.5-VL-72B\cite{qwen25vl}   & 768&      122.4s     & 77.9 & - &   -    &  71.2/77.8  \\
    Qwen2.5-VL-7B\cite{qwen25vl}   &  512 & 24.3s & 66.7 & 55.6          & - &63.3/69.7  \\
    InternVL-2.5-8B\cite{internvl2.5}   & 64 & 12.4s & -   & 60.0 & 68.9 &64.2/66.9   \\
    InternVideo-2.5-8B\cite{wang2025internvideo25}   &   512 &   33.8s &  63.9   & 60.6 & 72.8  &  65.1/-  \\
\midrule

\textbf{VideoChat-A1}~\footnotesize{(\textbf{\texttt{Qwen2.5-VL-7B}})}  &  42.0   & 18.6s   & 70.7 &   64.2  & 71.9       &  71.8/76.5  \\

\textbf{VideoChat-A1}~\footnotesize{(\textbf{\texttt{InternVL2.5-8B}})} &35.9 &  14.7s & 72.1 & 65.2  & 75.1   & 72.3/77.0 \\

\textbf{VideoChat-A1}~\footnotesize{(\textbf{\texttt{InternVideo2.5-8B}})}&  41.3  & 28.4s   &   70.1 & 65.4 & 76.2 &   72.9/76.8  \\

                \bottomrule
            \end{tabular}
        }
        }
        \caption{Comparison on Average Frame Number and Inference Time on Datasets. Inference Time includes the average time for video frame extraction, clustering, and the time required for the MLLM to answer all queries. 
        }
        \label{tab:efficient}
        \end{table*}

\endgroup

%% file: table/abla_global.tex
\begin{table}[!t]
  \centering
  \small
  \renewcommand{\arraystretch}{0.9} 
  \setlength{\tabcolsep}{3pt}        
  \resizebox{\columnwidth}{!}{%
    \begin{tabular}{c|c|cccc}
      \toprule
      \textbf{7/8B Agent} 
        & \begin{tabular}[c]{@{}c@{}}\textbf{Video}\\\textbf{Glance}\end{tabular}  
        & \textbf{Ego} & \textbf{LVBench} & \textbf{MLVU} & \textbf{VideoMME} \\
      \midrule
      \multirow{2}{*}{%
        \begin{tabular}[c]{@{}c@{}}
          \textbf{VideoChat-A1}\\
          \texttt{(Qwen2.5-VL-7B)}
        \end{tabular}
      }
        & \XSolidBrush & 69.4 & 63.4 & 70.0 & 70.4/74.9 \\
        & \Checkmark   & 70.7 & 64.2 & 71.9 & 71.8/76.5 \\
      \midrule
      \multirow{2}{*}{%
        \begin{tabular}[c]{@{}c@{}}
          \textbf{VideoChat-A1}\\
          \texttt{(InternVL2.5-8B)}
        \end{tabular}
      }
        & \XSolidBrush & 70.8 & 64.0 & 73.9 & 70.6/75.5 \\
        & \Checkmark   & 72.1 & 65.2 & 75.1 & 72.3/77.0 \\
      \midrule
      \multirow{2}{*}{%
        \begin{tabular}[c]{@{}c@{}}
          \textbf{VideoChat-A1}\\
          \texttt{(InternVideo2.5-8B)}
        \end{tabular}
      }
        & \XSolidBrush & 69.3 & 64.3 & 74.5 & 71.8/74.9 \\
        & \Checkmark   & 70.1 & 65.4 & 76.2 & 72.9/76.8 \\
      \bottomrule
    \end{tabular}
  }
  \caption{Ablation on the  Video Glance in \textbf{VideoChat-A1}.}
  \label{tab:abla_global_obv}
\end{table}

%% file: table/abla_reflect.tex
\begin{table}[!t]
  \centering
  \small
  \renewcommand{\arraystretch}{1.1}
  \resizebox{\columnwidth}{!}{%
    \begin{tabular}{cc|cccc}
      \toprule
      \textbf{SP} & \textbf{CoS} & \textbf{Ego} & \textbf{LVBench} & \textbf{MLVU} & \textbf{VideoMME} \\
      \midrule
      \XSolidBrush & \XSolidBrush & 67.1 & 56.4 & 66.2 & 63.4/69.8 \\
      \Checkmark   & \XSolidBrush & 67.5 & 61.1 & 68.5 & 67.8/73.9 \\
      \XSolidBrush & \Checkmark   & 69.1 & 57.7 & 67.0 & 65.8/70.8 \\
      \rowcolor{gray!20}
      \Checkmark   & \Checkmark   & \textbf{70.7} & \textbf{64.2} & \textbf{71.9} & \textbf{71.8/76.5} \\
      \bottomrule
    \end{tabular}
  }
  \caption{Ablation of Shot Partition (SP) and Chain-of-Shot (CoS) in \textbf{VideoChat-A1} (\texttt{Qwen2.5-VL-7B}).}
  \label{tab:process}
\end{table}

\begin{table}[!t]
  \centering
  \small
  \renewcommand{\arraystretch}{1.2}
  \resizebox{\columnwidth}{!}{%
    \begin{tabular}{c|cccc}
      \toprule
      \begin{tabular}[c]{@{}c@{}}\textbf{Method of}\\\textbf{Shot Partition}\end{tabular}
        & \textbf{Ego} & \textbf{LVBench} & \textbf{MLVU} & \textbf{VideoMME} \\
      \midrule
      Cluster Partition  & 68.3 & 58.4 & 67.7 & 67.8/72.8 \\
      Average Partition  & 69.1 & 57.7 & 67.0 & 65.8/70.8 \\
      \rowcolor{gray!20}
      Shot Partition     & \textbf{70.7} & \textbf{64.2} & \textbf{71.9} & \textbf{71.8/76.5} \\
      \bottomrule
    \end{tabular}
  }
  \caption{Ablation of different methods of shot partition in \textbf{VideoChat-A1} (\texttt{Qwen2.5-VL-7B}).}
  \label{tab:shot_partition}
\end{table}

%% file: table/abla_big.tex
\begin{table}[!t]
  \centering
  \small
  \renewcommand{\arraystretch}{0.9}
  \setlength{\tabcolsep}{3pt}
  \resizebox{\columnwidth}{!}{%
    \begin{tabular}{c|cccc}
      \toprule
      \textbf{Max Rounds} 
        & \textbf{Ego} & \textbf{LVBench}
        & \textbf{MLVU} & \textbf{VideoMME} \\
      \midrule
       1 & 67.3 & 60.1 & 66.2      & 66.8/73.4    \\
       2 & 70.0 & 62.5 & 69.7      & 69.6/76.0    \\
       3 & 70.7 & 64.2 & 71.9 & 71.8/76.5 \\
       4 & 70.9 & 64.5 & 72.1      & 72.0/76.9    \\
      \bottomrule
    \end{tabular}
  }
  \caption{Ablation over maximum iteration rounds in \textbf{VideoChat-A1} (\texttt{Qwen2.5-VL-7B}).}
  \label{tab:abla_maxround}
\end{table}

\begin{table}[!t]
  \centering
  \small
  \renewcommand{\arraystretch}{0.9}
  \resizebox{\columnwidth}{!}{%
    \begin{tabular}{c|cccc}
      \toprule
      \begin{tabular}[c]{@{}c@{}}\textbf{Number of}\\\textbf{Init Shots}\end{tabular}
        & \textbf{Ego} & \textbf{LVBench}
        & \textbf{MLVU} & \textbf{VideoMME} \\
      \midrule
       3 & 70.1 & 63.9 & 70.8      & 71.2/75.7    \\
      \rowcolor{gray!20}
       6 & \textbf{70.7} & \textbf{64.2} & \textbf{71.9} & \textbf{71.8/76.5} \\
       9 & 69.3 & 64.0 & 71.8      & 71.5/76.2    \\
      \bottomrule
    \end{tabular}
  }
  \caption{Ablation over number of initialized shots in \textbf{VideoChat-A1} (\texttt{Qwen2.5-VL-7B}).}
  \label{tab:num_init_shot}
\end{table}